%% file: ijcai23.tex
\documentclass{article}
\usepackage{ijcai23}

\usepackage{times}
\usepackage{soul}
\usepackage{url}
\usepackage[hidelinks]{hyperref}
\usepackage[utf8]{inputenc}
\usepackage[small]{caption}
\usepackage{graphicx}
\usepackage{amsmath}
\usepackage{amsthm}
\usepackage{booktabs}
\usepackage{algorithm}
\usepackage{algorithmic}
\usepackage[switch]{lineno}

\usepackage{etoolbox}
\apptocmd{\thebibliography}{\setlength{\itemsep}{0pt}}{}{}

\usepackage{makecell}
\usepackage{pifont}
\usepackage{amssymb}
\usepackage{color}
\usepackage{tikz}
\usepackage[edges]{forest}
\definecolor{hidden-draw}{RGB}{205, 44, 36}
\definecolor{hidden-blue}{RGB}{194,232,247}
\definecolor{hidden-orange}{RGB}{243,202,120}
\definecolor{hidden-yellow}{RGB}{242,244,193}

\title{Generalizing to Unseen Elements: A Survey on Knowledge Extrapolation for Knowledge Graphs}

\author{
Mingyang Chen$^1$ \and
Wen Zhang$^2$ \and
Yuxia Geng$^1$ \and
Zezhong Xu$^1$ \and
Jeff Z. Pan$^3$ \and
Huajun Chen$^{1,4,5}$\footnote{Corresponding author.}
\affiliations
$^1$College of Computer Science and Technology, Zhejiang University\\
$^2$School of Software Technology, Zhejiang University\\
$^3$School of Informatics, The University of Edinburgh\\
$^4$Donghai Laboratory \\
$^5$Alibaba-Zhejiang University Joint Institute of Frontier Technologies \\
\emails
\{mingyangchen, zhang.wen, gengyx, xuzezhong, huajunsir\}@zju.edu.cn,
j.z.pan@ed.ac.uk
}

\begin{document}

\maketitle

\begin{abstract}
Knowledge graphs (KGs) have become valuable knowledge resources in various applications, and knowledge graph embedding (KGE) methods have garnered increasing attention in recent years. However, conventional KGE methods still face challenges when it comes to handling unseen entities or relations during model testing. To address this issue, much effort has been devoted to various fields of KGs.
In this paper, we use a set of general terminologies to unify these methods and refer to them collectively as \textit{Knowledge Extrapolation}. We comprehensively summarize these methods, classified by our proposed taxonomy, and describe their interrelationships. Additionally, we introduce benchmarks and provide comparisons of these methods based on aspects that are not captured by the taxonomy. Finally, we suggest potential directions for future research.

\end{abstract}

\section{Introduction}

A knowledge graph (KG)~\cite{PVGW2017} comprises triples in the form of facts, such as \textit{(head entity, relation, tail entity)}, with entities and relations represented as nodes and edges in a graph. Despite their simplicity, KGs are increasingly recognized as valuable knowledge resources for various applications.

Knowledge graph embedding (KGE) is a dominant field that has recently gained massive attention for representing elements (i.e., entities and relations) of knowledge graphs in latent vector spaces.
Conventional KGE methods are evaluated among entities and relations occurring in triples used for model training, assuming a fixed entity and relation set. However, this assumption does not always hold in real-world applications \cite{KnowTransOOKB-IJCAI2017,GMatching-EMNLP2018}.

Knowledge graphs are dynamic, with new entities and relations emerging over time. Furthermore, the number of knowledge graphs is increasing.
Conventional KGE methods are incapable of handling emerging elements of KGs because they cannot map unseen elements to a proper position in the vector space of trained elements. This inability to generalize to unseen elements limits their usefulness in addressing the evolution feature of KGs.

Recently, many studies have focused on generalizing to unseen elements of KGs in various scenarios. For instance, some research has concentrated on predicting missing triples for out-of-knowledge-base (\textit{OOKB}) entities \cite{KnowTransOOKB-IJCAI2017,LAN-AAAI2019}. Additionally, some \textit{inductive} relation prediction methods and toolkits have studied generalizing to entirely new KGs with unseen entities \cite{GraIL-ICML2020,MorsE-SIGIR2022,NeuralKG-ind}. Furthermore, problems of generalizing to unseen relations have also been deeply investigated, especially in low-resource settings such as \textit{few-shot} \cite{GMatching-EMNLP2018,MetaR-EMNLP2019} and \textit{zero-shot} \cite{OntoZSL-WWW2021} scenarios.

Therefore, the problem of generalizing to unseen elements has garnered increasing attention in the knowledge graph field. However, investigations of this problem have used different scenarios and terminologies, such as OOKB, inductive, few-shot, and zero-shot.
Furthermore, there is no existing literature that systematically summarizes this area.
Based on our research in this field, we ask the following question: \textit{Can we use a general set of terminologies to comprehensively summarize works in this area and make comparisons to provide insights for future research?}
In this paper, we present the first survey of recent studies on generalizing to unseen elements in KGs and refer to them collectively as \textit{Knowledge Extrapolation}.
Specifically, we begin by providing background on knowledge extrapolation and unifying related definitions.
Next, we offer an overview of knowledge extrapolation methods and categorize them using our proposed taxonomy.
Additionally, we summarize commonly used benchmarks, compare methods, and suggest directions for future research.

\section{Preliminary}

\subsection{Knowledge Graph Embedding}
To begin, we provide a formal definition of knowledge graphs. Knowledge graphs are commonly represented as graph structures, in which nodes represent entities corresponding to specific concepts, and edges are labeled by relations.
Formally, a knowledge graph can be denoted as $\mathcal{G} = (\mathcal{E}, \mathcal{R}, \mathcal{T})$, where $\mathcal{E}$ is the set of entities, $\mathcal{R}$ is the set of relations, and $\mathcal{T}$ is the set of triples.
A triple describes a connection between two entities through a relation, and can be represented as $(h, r, t) \subseteq \mathcal{E} \times \mathcal{R} \times \mathcal{E}$.
In this context, both entities and relations are considered elements within a KG, as is common in the RDF~\cite{Pan2009} format.

Based on knowledge graphs, the goal of conventional knowledge graph embedding methods is to embed elements in $\mathcal{E}$ and $\mathcal{R}$ into continuous vector spaces while preserving the inherent structure of a KG \cite{kgembedding}. 
Link prediction is a common task for evaluating the effectiveness of a KGE method, and we use it as an example to introduce the basics since most existing works focus on it. However, exploring KGE and knowledge extrapolation on other KG-related tasks is a pressing need (cf. \S \ref{sec:diver_application}).

In practice, as shown in Figure \ref{fig:setting}, there is a training set $\mathcal{G}^{tr} = (\mathcal{E}^{tr}, \mathcal{R}^{tr}, \mathcal{T}^{tr})$ (a) and a test set $\mathcal{G}^{te} = (\mathcal{E}^{te}, \mathcal{R}^{te}, \mathcal{T}^{te})$ (b).
The goal of link prediction is to train the model to score positive triples higher than negative triples and generalize the model to the test set. This can be expressed as $\min_{\theta} \mathbb{E}_{(h, r, t)\in \mathcal{T}^{te}}[\ell_{\theta}(h, r, t)]$, where $\ell_{\theta}(h, r, t) \propto -s(\mathbf{h}, \mathbf{r}, \mathbf{t})+s(\mathbf{h}^{\prime}, \mathbf{r}^{\prime}, \mathbf{t}^{\prime})$ is the loss function, $s$ is the score function, and $\theta$ represents the learnable parameters, including entity and relation embeddings.
We use $\mathbf{h}$, $\mathbf{r}$, $\mathbf{t}$ to denote the embeddings of $h$, $r$, $t$ and $(h^{\prime},r^{\prime},t^{\prime}) \notin \mathcal{T}^{tr}$ to denote negative samples.

However, conventional KGE methods cannot deal with new entities and relations during the test since they don't have proper embeddings for new elements. Knowledge extrapolation focuses on solving this problem, and we describe the details next.

\subsection{Knowledge Extrapolation Settings}

Methods aimed at knowledge extrapolation attempt to conduct link prediction on unseen elements. To unify existing works on handling these unseen elements, we introduce a set of general terms.
Specifically, during knowledge extrapolation, there are two sets used for testing: one provides support information about the unseen elements (such as their structural or textual features), and the other evaluates the model's link prediction ability, much like the original $\mathcal{T}^{te}$.
We refer to these sets as the support set $\mathcal{S}^{te}$ and query set $\mathcal{Q}^{te}$, respectively, and the test set is formulated as $\mathcal{G}^{te} = (\mathcal{E}^{te}, \mathcal{R}^{te}, \mathcal{S}^{te}, \mathcal{Q}^{te})$.
While different works may use varying terminology, they must all involve these two sets during knowledge extrapolation. For convenience, we refer to them uniformly as the support and query sets.

\begin{figure}[t]
\centering
\includegraphics[scale=0.7]{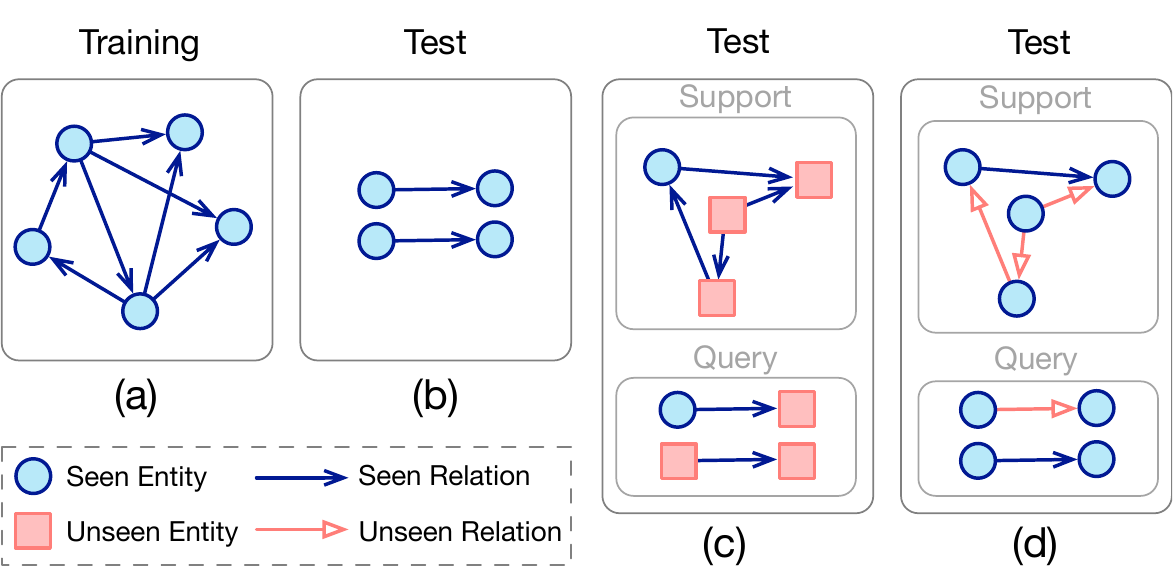}
\caption{Training (a) and test (b) set for conventional knowledge graph embedding. Example test set for the entity extrapolation setting (c) and the relation extrapolation setting (d). Note that there may be any support information about unseen elements in the support set, and we use relevant triples as examples.}
\label{fig:setting}
\end{figure}

In this work, we categorize existing approaches for handling unseen elements into two types: \textit{Entity Extrapolation} and \textit{Relation Extrapolation}.
As illustrated in Figure \ref{fig:setting}, we use the term ``entity extrapolation" to refer to situations where previously unseen entities appear in the test set, and ``relation extrapolation" to describe scenarios where previously unseen relations are present in the test set.

Recent years have seen fast growth in knowledge extrapolation for KGs. Early work on rule learning \cite{AMIE-WWW2013,RuleN-ISWC2018} addressed entity extrapolation via rules that are independent of specific entities. Later, researchers \cite{DKRL-AAAI2016,OWE-AAAI2019} used textual descriptions to handle unseen entities. With GNNs becoming popular for KGs, methods \cite{KnowTransOOKB-IJCAI2017,LAN-AAAI2019} investigated aggregating knowledge from seen entities to unseen entities. GraIL \cite{GraIL-ICML2020} used subgraph encoding to handle unseen entities in a completely new KG, with many methods following this paradigm. Other works \cite{MorsE-SIGIR2022,NodePiece-ICLR2022} explored encoding entity-independent information for unseen entities. Meanwhile, some models \cite{KEPLER-TACL2019,StAR-WWW2021} incorporated pre-trained language models to encode textual descriptions for unseen entities. Recent research has also considered few-shot settings for handling unseen entities \cite{GEN-NIPS2020,HRFN-CIKM2021}. For relation extrapolation, pioneer works \cite{GMatching-EMNLP2018,MetaR-EMNLP2019} have often focused on few-shot settings. Other works represent unseen relations using textual descriptions \cite{ZSGAN-AAAI2020} or ontologies \cite{OntoZSL-WWW2021}, without using support triples, referred to as zero-shot learning for KGs.

\input{category.tex}

\section{Methods}
In this section, we describe the existing works on knowledge extrapolation in detail. As shown in Figure \ref{fig:texonomy}, we categorize these methods based on their model design. 
For each category of methods, we begin by introducing its general idea and then delve into the specifics of existing methods.

\subsection{Entity Extrapolation}
\label{sec:entity-ext}
\subsubsection{Entity Encoding-based Entity Extrapolation}
Conventional knowledge graph embedding methods typically learn an embedding lookup table for entities. However, this paradigm hinders models from extrapolating to unseen entities, as there are no reasonable embeddings for them in the learned table.
To handle unseen entities, an intuitive approach is to learn to encode entities rather than learn fixed embedding tables. These learned encoders can operate on the support set of entities to produce reasonable embeddings for them.
The score function of entity encoding-based entity extrapolation for $(h,r,t) \in \mathcal{Q}^{te}$ is formulated as:
\begin{equation}
    s(f(h, \mathcal{S}^{te}),\mathbf{r},f(t, \mathcal{S}^{te}))
\label{eq:ee-ee}
\end{equation}
where $f$ denotes the learnable encoder, and $\mathbf{r}$ denotes relation embedding or relation-dependent parameters. Note that not all methods encode both head and tail entities.

Existing works have designed various encoding models $f$ that correspond to different types of information in the support set $\mathcal{S}^{te}$.
If the support set contains triples about unseen entities, then $f$ encodes those entities \textit{from structural information}.
On the other hand, if the support set contains other types of information (e.g., textual descriptions) about unseen entities, we refer to this situation as encoding unseen entities \textit{from other information}.

\paragraph{{Encode from structural information.}}
Some methods assume that unseen entities are connected to seen entities and focus on transferring knowledge from the latter to the former by aggregating neighbors for the unseen entities.
For example, Hamaguchi \textit{et al.}~\shortcite{KnowTransOOKB-IJCAI2017} apply
a graph neural network to propagate representations between entities and represent unseen emerging entities using different transition and pooling functions.
To build more efficient neighbor aggregators for emerging entities, LAN \cite{LAN-AAAI2019} uses attention weights based on logic rules and neighbor entity embeddings that consider query relations.
VN Network \cite{VN-network-CIKM2020} conducts virtual neighbor prediction based on logic and symmetric path rules to make the neighbor for an unseen entity denser.
Bhowmik and de Melo~\shortcite{Explainable-ISWC2020} propose a graph transformer architecture to aggregate the neighbor information for learning entity embeddings and use the policy gradient to find a symbolic reasoning path for explainable reasoning.
Albooyeh \textit{et al.}~\shortcite{OOS-EMNLP2020} designs a training procedure resembling what is expected of the model at the test time.  
CFAG \cite{CFAG-AAAI2022} first uses an aggregator to obtain entity representations and then applies a GAN to generate entity representations based on the query relation.
ARGCN \cite{ARGCN-CIKM2022} considers representing new entities emerging in multiple batches.
QBLP \cite{QBLP-ISWC2021} explores the benefits of employing hyper-relational KGs on entity extrapolation.

In addition to considering connections between seen and unseen entities, recent works have also started to explore relations between unseen entities.
GEN \cite{GEN-NIPS2020} considers unseen entities with only a few associative triples and tackles link prediction between seen to unseen entities, as well as between unseen entities, by designing a meta-learning framework to conduct learning to extrapolate the knowledge from a given KG to unseen entities.
Following GEN, HRFN \cite{HRFN-CIKM2021} first coarsely represents unseen entities by meta-learned hyper-relation features and then uses a GNN to obtain fine-grained embeddings for unseen entities. 

In certain scenarios, unseen entities have no connection with seen entities and form a completely new KG. Consequently, some methods assume that the training and test entity sets are not overlapping, i.e., $\mathcal{E}^{te} \cap \mathcal{E}^{tr} = \emptyset$. For these scenarios, the methods do not focus on aggregating representations from seen entities for unseen entities. Instead, they learn to encode entity-independent information to output entity embeddings.
MorsE \cite{MorsE-SIGIR2022} uses an entity initializer and a GNN modulator to model entity-independent meta-knowledge to produce high-quality embeddings for unseen entities and learn such meta-knowledge by meta-learning.
NodePiece \cite{NodePiece-ICLR2022} also treats connected relations as entity-independent information and tries MLP and Transformer to encode relational contexts for unseen entities.
Unlike the above methods encoding entities individually, INDIGO \cite{INDIGO-NIPS2021} encodes the support triples and query candidate triples into a node-annotated graph where nodes correspond to pairs of entities, and uses GNN to update node features for indicating the query triple predictions.

\paragraph{Encode from other information.} 
In addition to using relevant triples as support sets for unseen entities, other types of information can also be useful for encoding unseen entities. The textual description is common for encoding unseen entities.
DKRL \cite{DKRL-AAAI2016} embodies entity descriptions for knowledge graph embedding and jointly trains structure-based embeddings and description-based embeddings.
ConMask \cite{ConMask-AAAI2018} applies relationship-dependent content masking on a given entity description and uses the representation extracted from the description to predict the most proper entities from an existing entity set.
OWE \cite{OWE-AAAI2019} trains traditional knowledge graph embeddings and word embeddings independently, and learns a transformation from entities' textual information to their embeddings, enabling unseen entities with text can be mapped into the KGE space.
KEPLER \cite{KEPLER-TACL2019} uses the same pre-trained language model to encode entities and texts, and jointly optimizes it with knowledge embedding and masked language modeling objectives.
StAR \cite{StAR-WWW2021} is a structure-augmented text representation model which encodes triples with a siamese-style textual encoder. Since it encodes the text of entities, it can easily extrapolate to unseen entities using their text.
BLP \cite{BLP-WWW2021} treats a pre-trained language model as an entity encoder and finetunes it via link prediction.
SimKGC \cite{SimKGC-ACL2022} uses contrastive learning with different negative samplings to improve the efficiency of text-based knowledge graph embedding. 
StATIK \cite{StATIK-NAACL2022} combines a language model and a message passing graph neural network as an encoder to process and uses TransE as the decoder to score triples.
 
\subsubsection{Subgraph Predicting-based Entity Extrapolation}
The aforementioned entity encoding-based methods commonly treat the head entity, relation, and tail entity in a triple individually.  
However, some research provides another view that does not explicitly encode entities. Instead, they treat the head and tail entity in a triple together and encode the relational subgraph between them. This perspective assumes that the semantics underlying the subgraph between two entities can be used to predict their relation.
The ability to encode the subgraph of two entities can be extrapolated to unseen entities since subgraph structures are independent of entities. 
For the subgraph predicting-based method, the score function can be formulated as:
\begin{equation}
    s(f(h, t, \mathcal{S}^{te}),\mathbf{r})
\label{eq:ee-rse}
\end{equation}
where $f$ is responsible for subgraph extraction and encoding; $\mathbf{r}$ denotes relation embedding or relation-dependent parameters; $s$ is responsible for predicting whether the subgraph is related to a specific relation.
This research line mainly encodes subgraphs for unseen entities based on triples from support sets, so they assume that support sets contain structural information about unseen entities.

One line of this method starts from GraIL \cite{GraIL-ICML2020}, which uses a GNN to reason over enclosing subgraph between target nodes and learns entity-independent relational semantics for entity extrapolation. 
Specifically, for predicting the relation between two unseen entities, GraIL first extracts the enclosing subgraph around them, then labels each entity in this subgraph based on the distance between that entity and two unseen entities. Finally, it uses a GNN to score the subgraph with candidate relations. 
CoMPILE \cite{CoMPILE-AAAI2021} proposes a node-edge communicative message-passing mechanism replacing the original GNN to consider the importance of relations.
TACT \cite{TACT-AAAI2021} considers correlations between relations in a subgraph, and encodes a Relational Correlation Network (RCN) to enhance the encoding of the enclosing subgraph.
ConGLR \cite{ConGLR-SIGIR2022} formulates a context graph representing relational paths from a subgraph and then uses two GCNs to process it with the enclosing subgraph respectively.
Since extracted enclosing subgraphs can be sparse and some surrounded relations are neglected, SNRI \cite{SNRI-IJCAI2022} leverages complete neighbor relations of entities in a subgraph by neighboring relational features and neighboring relational paths.
BertRL \cite{BertRL-AAAI2022} leverages pre-trained language models to encode reasoning paths to enhance subgraph semantics.
RMPI \cite{RMPI-ICDE2023} utilizes the relation semantics defined in the KG’s
ontological schema~\cite{WPKD2020} to extend 
for handling both unseen entities and relations.

The path between two entities can also be viewed as a subgraph
and used to indicate the relation between them.
PathCon \cite{PathCon-KDD2021} proposes a relational message passing framework and conducts it to encode relational contexts and relational paths for two given entities to predict the missing relation.
NBFNet \cite{NBFNet-NIPS2021} uses the Message and Aggregate operation from GNN to parameterize the generalized Bellman-Ford algorithm for representing the paths between two entities.
RED-GNN \cite{RED-GNN-WWW2022} proposes a relational directed graph (r-digraph) to preserve the local information between entities and recursively constructs the r-digraph between one query and any candidate entity, making the reasoning processes efficient.

\subsubsection{Rule Learning-based Entity Extrapolation}
Several works have explored learning rules from knowledge graphs, as these logical rules can inherently extrapolate to unseen entities since they are independent of specific entities.
Rule learning-based methods can be divided into two categories. The pure symbolic methods generate rules from existing knowledge through statistics and filter them with predefined indicators. AMIE~\cite{AMIE-WWW2013} proposes predefined indicators, including head coverage and confidence, to filter possible rules which are generated with three different atom operators. 
RuleN~\cite{RuleN-ISWC2018} and AnyBURL~\cite{AnyBURL-IJCAI2019} generate possible rules by sampling the path from the head entity to the tail entity of each triple and filtering them using confidence and head coverage. 
Another kind of method combines neural networks and symbolic rules. Neural LP~\cite{NeuralLP-NIPS2017} and DRUM~\cite{DRUM-NIPS2018} are based on TensorLog and use neural networks to simulate rule paths. 
CBGNN \cite{CBGNN-ICML2022} view logical rules as cycles from the perspective of algebraic topology and learn rules in the space of cycles.

\subsection{Relation Extrapolation}
\subsubsection{Relation Encoding-based Relation Extrapolation}
Similar to entity extrapolation, the shortcoming of conventional knowledge graph embedding methods on relation extrapolation is that they cannot give unseen relations reasonable embeddings. However, since some observed information from the support set for unseen relations can be utilized, encoding such information to embed unseen relations is an intuitive solution. The score function of relation encoding-based methods is formulated as follows: 
\begin{equation}
    s(\mathbf{h}, f(r, \mathcal{S}^{te}), \mathbf{t})
\label{eq:re-re}
\end{equation}
where $f$ is an encoder that transforms related support information for an unseen relation $r$ to its embedding, and $\mathbf{h}$ and $\mathbf{t}$ are entity embeddings or entity-dependent parameters. 
Depending on the type of information used for relation encoding, we also categorize these methods into \textit{encode from structural information} and \textit{encode from other information}.

\paragraph{Encode from structural information.}
Methods in this category assume that there are some example triples (i.e., structural information) for unseen relations in the support set, and the connected head and tail entities of unseen relations can reveal their semantics. These methods mainly focus on transferring representation from entities to unseen relations.
MetaR \cite{MetaR-EMNLP2019} learns relation-specific meta information from support triples to replace the missing representations for unseen relations. More precisely, for a specific unseen relation, MetaR first encodes entity pairs of it and outputs relation meta. Then, a rapid gradient descent step is applied to the relation meta following optimization-based meta-learning, which will produce good generalization performance on that unseen relation.
In order to leverage more structural information provided by connected entities of unseen relations, GANA \cite{GANA-SIGIR2021} uses a gated and attentive neighbor aggregator to represent entity representations. 
Such entity representations are used to generate general representations for unseen relations via an attentive Bi-LSTM encoder. Furthermore, it applies the training paradigm of optimization-based meta-learning with TransH as the score function.

\paragraph{Encode from other information.}
In addition to using structural information from relevant triples, methods in this category rely on unseen relations' support information from external resources, including textual descriptions and ontological schemas.
ZSGAN \cite{ZSGAN-AAAI2020} leverages Generative Adversarial Networks (GANs) to generate plausible relation embeddings for unseen relations conditioned on their encoded textual descriptions. 
Since a KG is often accompanied by an ontology as its schema, OntoZSL \cite{OntoZSL-WWW2021} is proposed to synthesize the unseen relation embeddings conditioned on the embeddings of an ontological schema which includes the semantics of the KG relations.
DOZSL \cite{DOZSL-KDD2022} proposes a property-guided disentangled ontology embedding method to extract more fine-grained inter-class relationships between seen and unseen relations in the ontology. It uses a GAN-based
generative model and a GCN-based propagation model to integrate disentangled embeddings for generating embeddings for unseen relations.
Besides leveraging generative models such as GANs to obtain plausible relation embeddings from support information, DMoG \cite{DMoG-COLING2022} learns a linear mapping matrix to transform the encoded textual and ontological features, and HAPZSL \cite{HAPZSL-NeuroComp2022} uses an attention mechanism to encode relation descriptions into valid relation prototype representations.

\subsubsection{Entity Pair Matching-based Relation Extrapolation}
Another solution, instead of encoding the relation directly, is to encode head and tail entity pairs of an unseen relation and then match these encoded entity pairs with entity pairs in the query set to predict whether they are connected by the same unseen relation.
The score function of entity pair matching-based methods for $(h,r,t)$ can be formulated as follows:
\begin{equation}
\begin{aligned}
    & s(f(\operatorname{PAIR}(r, \mathcal{S}^{te})), f(h, t)) \\
    & \operatorname{PAIR}(r, \mathcal{S}^{te}) = \{(h_{i}, t_{i}) | (h_i, r, t_i) \in \mathcal{S}^{te} \}
\label{eq:re-epe}
\end{aligned}
\end{equation}
where $\operatorname{PAIR}(r, \mathcal{S}^{te})$ is used to get the head-tail entity pairs from the support set for the unseen relation.
$f$ encodes head-tail entity pairs, and $s$ scores for entity pair matching. Since this research line involves encoding entity pairs for unseen relations, the support sets are commonly assumed to consist of relevant triples for these unseen relations.

GMatching \cite{GMatching-EMNLP2018} uses entity embeddings and entity local graph structures to represent entity pairs and learn a metric model to discover more similar triples. 
FSRL \cite{FSRL-AAAI2020} uses a relation-aware heterogeneous neighbor encoder considering different impacts of neighbors of an entity, and a recurrent
autoencoder aggregation network to aggregate multiple reference entity pairs in the support set.
FAAN \cite{FAAN-EMNLP2020} uses an adaptive neighbor encoder to encode entities and uses a Transformer to encode entity pairs; furthermore, an adaptive matching processor is introduced to match a query entity pair to multiple support entity pairs attentively.
MetaP \cite{MetaP-SIGIR2021} learns relation-specific meta patterns from entity pairs of a relation, and such patterns are captured by convolutional filters on representations (by pre-trained or random initialize) of entity pairs. 
Furthermore, the subgraphs between the entity pairs are also expressive. 
P-INT \cite{P-INT-EMNLP2021} leverages the paths between an entity pair to represent it and calculates the path similarity between support entity pairs and query entity pairs to predict the query triples about unseen relations.
GraphANGEL \cite{GraphANGEL-ICLR2022} extracts 3-cycle and 4-cycle patterns to represent an unseen relation; specifically, for a query triple with an unseen relation, it extracts target patterns, supporting patterns, refuting patterns for query entity pairs and calculates the similarity between target patterns and support patterns as well refuting patterns. 
CSR \cite{CSR-NIPS2022} uses connection subgraphs to represent entity pairs. It finds the shared connection subgraph among the support triples and tests whether it connects query triples.

\section{Benchmarks}
\label{sec:benchmarks}

\begin{table*}[t]
\centering
\resizebox{\textwidth}{!}{
\begin{tabular}{ccll}
\toprule
Dataset & Type & Proposed by & Used by \\
\midrule
\ding{172} & Ent & MEAN \cite{KnowTransOOKB-IJCAI2017} & 
\makecell[l]{
MEAN \cite{KnowTransOOKB-IJCAI2017}, LAN \cite{LAN-AAAI2019}, VN Network \\ \cite{VN-network-CIKM2020}, INDIGO \cite{INDIGO-NIPS2021} 
}\\
\midrule
\ding{173} & Ent & GraIL \cite{GraIL-ICML2020} & \makecell[l]{
GraIL \cite{GraIL-ICML2020}, CoMPILE \cite{CoMPILE-AAAI2021}, TACT \cite{TACT-AAAI2021}, \\ ConGLR \cite{ConGLR-SIGIR2022}, SNRI \cite{SNRI-IJCAI2022}, BertRL \cite{BertRL-AAAI2022}, \\ RMPI \cite{RMPI-ICDE2023}, MorsE \cite{MorsE-SIGIR2022}, NodePiece \\ \cite{NodePiece-ICLR2022}, NBFNet \cite{NBFNet-NIPS2021}, RED-GNN \cite{RED-GNN-WWW2022}, \\ INDIGO \cite{INDIGO-NIPS2021}, RuleN \cite{RuleN-ISWC2018}, Nerual LP \\ \cite{NeuralLP-NIPS2017}, DRUM \cite{DRUM-NIPS2018}, CBGNN \cite{CBGNN-ICML2022}
}\\ 
\midrule
\ding{174} & Rel & GMatching \cite{GMatching-EMNLP2018} & 
\makecell[l]{
GMatching \cite{GMatching-EMNLP2018}, MetaR \cite{MetaR-EMNLP2019}, FSRL \\ \cite{FSRL-AAAI2020}, FAAN \cite{FAAN-EMNLP2020}, GANA \cite{GANA-SIGIR2021}, \\ MetaP \cite{MetaP-SIGIR2021}, P-INT \cite{P-INT-EMNLP2021}, CSR \cite{CSR-NIPS2022}
}\\
\midrule
\ding{175} & Rel & ZSGAN \cite{ZSGAN-AAAI2020} & \makecell[l]{ZSGAN \cite{ZSGAN-AAAI2020}, OntoZSL \cite{OntoZSL-WWW2021}, DOZSL \cite{DOZSL-KDD2022} \\ DMoG \cite{DMoG-COLING2022}, HAPZSL \cite{HAPZSL-NeuroComp2022}} \\
\bottomrule
\end{tabular}
}
\caption{The usage of four commonly used benchmarks described in \S \ref{sec:benchmarks}. The column `Type' indicates the dataset is used for entity or relation extrapolation.}
\label{tab:benchmarks}
\end{table*}

In this section, we briefly describe some datasets used to evaluate models for knowledge extrapolation. The following four datasets are commonly used by different knowledge extrapolation models, and they have various assumptions for diverse application scenarios.
We show the usage details in Table \ref{tab:benchmarks} and describe them as follows.
\paragraph{\ding{172} WN11-\{Head/Tail/Both\}-\{1,000/3,000/5,000\}} Here are nine datasets derived from WordNet11 created by Hamaguchi \textit{et al.}~\shortcite{KnowTransOOKB-IJCAI2017}. These datasets are used to conduct entity extrapolation experiments, and they assume that unseen entities are all connected to training entities since triples that contain two unseen entities in support sets are discarded. Head/Tail/Both denotes the position of unseen entities in test triples, and 1,000/3,000/5,000 represents the number of triplets used for generating the unseen entities.
\paragraph{\ding{173} \{WN18RR/FB15k-237/NELL995\}-\{v1/2/3/4\}} These 12 datasets are generated by Teru \textit{et al.}~\shortcite{GraIL-ICML2020} from WN18RR, FB15k-237, and NELL955. Each original dataset is sampled into four different versions with an increasing number of entities and relations. In these datasets, the entity sets during training and testing are disjoint, which indicates that they are used to evaluate entity extrapolation models assuming unseen entities are not connected to training entities.
\paragraph{\ding{174} NELL-One/Wiki-One} These two datasets are developed by Xiong \textit{et al.}~\shortcite{GMatching-EMNLP2018} and used to evaluate the few-shot relation prediction task originally. 
The relation sets for the training and the testing are disjoint, indicating models evaluated on these datasets are capable of relation extrapolation.
For each unseen relation during the test, the number of support triples (i.e., $k$) is often specified, and the task is called $k$-shot link prediction. 
\paragraph{\ding{175} NELL-ZS/Wiki-ZS} These two datasets are presented by Qin \textit{et al.}~\shortcite{ZSGAN-AAAI2020}, and `ZS' denotes that they are used to evaluate zero-shot learning for link prediction, i.e., relation extrapolation with information other than triples related to unseen relations. Each relation in these datasets has its textual description, and such information is viewed as support sets for unseen relations. Furthermore, besides textual information, Geng \textit{et al.}~\shortcite{OntoZSL-WWW2021} adds ontological schemas for these two datasets as side information for relations.

Moreover, there are many other datasets provided by different works to show their own effectiveness from some specific aspects. 
LAN \cite{LAN-AAAI2019} and VN Network \cite{VN-network-CIKM2020} use FB15k and YAGO37 to construct datasets, respectively. GEN \cite{GEN-NIPS2020} and HRFN \cite{HRFN-CIKM2021} develop datasets to demonstrate their link prediction ability on both seen and unseen entities.
MaKEr \cite{MaKEr-IJCAI2022} and RMPI \cite{RMPI-ICDE2023} construct datasets showing that they can handle both unseen entities and relations during the model test.
ARGCN \cite{ARGCN-CIKM2022} and LKGE \cite{LKGE-AAAI2023} create datasets to simulate the growth of knowledge graphs with emerging entities and relations.

\section{Comparison and Discussion}
\subsection{Assumption on Entity Extrapolation}

As described in \S \ref{sec:entity-ext}, there are two different assumptions about entity extrapolation. One assumption is that unseen entities in support sets are connected to seen entities (i.e., $\mathcal{E}^{te} \cap \mathcal{E}^{tr} \neq \emptyset$), while the other is that unseen entities form entirely new KGs in support sets and are not connected by seen entities (i.e., $\mathcal{E}^{te} \cap \mathcal{E}^{tr} = \emptyset$). 
We refer to these two assumptions as semi-entity extrapolation and fully-entity extrapolation for convenience. Generally, methods designed for fully-entity extrapolation can handle semi-entity extrapolation problems, but not vice versa.
We discuss the ability of entity extrapolation methods to handle these two assumptions as follows.

Most semi-entity extrapolation models lie in entity encoding-based methods and encode unseen entities from structural information since they often design modules for transferring knowledge from seen entities to unseen entities by aggregating representations from seen entities \cite{KnowTransOOKB-IJCAI2017,LAN-AAAI2019,CFAG-AAAI2022}. 
Some other methods that conduct entity encoding from structural information can also handle fully-entity extrapolation by designing entity-independent encoding procedure \cite{MorsE-SIGIR2022,NodePiece-ICLR2022}.

Furthermore, methods encoding unseen entities from other information (e.g., textual descriptions) can usually solve fully-entity extrapolation problems since such side information is sufficient for encoding entity embeddings and such encoding is independent to specific entities \cite{StAR-WWW2021,SimKGC-ACL2022}.
Subgraph predicting-based methods and rule learning-based methods are also capable of tackling fully-entity extrapolation since subgraphs and rules are entity-independent.

\subsection{Information in Support Set}
Various types of information have been explored to build support sets for unseen elements, including triples, textual descriptions, and ontologies. Here, we compare these three widely used types of support information as follows.

Triples, which provide structural information, are an intuitive type of support information for unseen elements since they usually emerge with other elements in the form of triples rather than alone. Knowledge from seen elements provided by triples can be utilized by unseen elements. 
For example, in entity extrapolation, surrounding seen relations can provide type information for unseen entities \cite{MorsE-SIGIR2022}; in relation extrapolation, connected seen entities can reveal the characteristics of unseen relations \cite{GMatching-EMNLP2018,MetaR-EMNLP2019}."

Textual descriptions are also common for KGs since many KGs are constructed from text data. Text descriptions can naturally provide the ability to extrapolate to unseen elements, and the typical procedure is to use text encoders (e.g., pre-trained language models) to transform text descriptions into embeddings. That is, the encoder $f$ in Eq. (\ref{eq:ee-ee}) or Eq. (\ref{eq:re-re}) can be treated as a text encoder. Both entity extrapolation \cite{StAR-WWW2021,BLP-WWW2021} and relation extrapolation \cite{ZSGAN-AAAI2020} can benefit from textual descriptions.

Ontologies are typically used as prior knowledge about the correlation between seen and unseen elements and have been widely used to handle unseen relations in existing works. An ontology is often represented as a graph encompassing relation hierarchies and constraints on relation domains and ranges. The embeddings of unseen relations can be generated using an ontology-based approach that utilizes various techniques, such as GANs \cite{OntoZSL-WWW2021} or disentangled representation learning \cite{DOZSL-KDD2022}.

Besides using the above support information alone, some methods apply different support information together. For example, BertRL \cite{BertRL-AAAI2022} uses both subgraph structures and textualized paths to predict relations between unseen entities, exploiting both structural triples and textual descriptions. Similarly, RMPI \cite{RMPI-ICDE2023} uses subgraphs and ontological schemas together to tackle both unseen entities and relations.

\section{Future Prospects}
\subsection{Diverse Application}
\label{sec:diver_application}
Most existing knowledge extrapolation methods are evaluated by simple link prediction on test sets. 
Even though the link prediction task can show the effectiveness of models and help knowledge graph completion, it is valuable to explore how to generalize to unseen KG elements on diverse applications.
GNN-QE \cite{GNN-QE-NIPS2022} studies answering logical queries expressed in a subset of first-order logic with unseen entities at test time.
Meanwhile, ContEA \cite{ContEA-ISWC2022} targets the entity alignment task under the growing of KGs.
We argue that besides the above in-KG applications, more common out-of-KG tasks like KG enhanced question answering \cite{MorsE-SIGIR2022}, including those based on LLMs~\cite{HWQM+2023} can be explored.

\subsection{Multi-modal Support Information}
Multi-modal knowledge graphs are a current research topic that has been explored in recent literature. While existing knowledge extrapolation methods have primarily focused on using natural language as support information for unseen elements, there are few works that address the potential of utilizing visual information. However, we believe that images can also help generalize KGE to unseen elements since they can be universally understood by specific pre-trained encoders. Additionally, hyper-relational KGs~\cite{QBLP-ISWC2021}, where triples can be instantiated with a set of qualifiers, can provide different modal information. 

\subsection{Entity and Relation Extrapolation}
Existing research on knowledge extrapolation primarily focuses on solving entity extrapolation and relation extrapolation separately, but in real-world applications, unseen entities and relations may emerge simultaneously.
One feasible solution is effectively integrating methods of entity extrapolation and relation extrapolation. Existing literature has made some attempts.
Specifically, MaKEr \cite{MaKEr-IJCAI2022} uses meta-learning to learn to encode unseen entities and relations based on a set of training tasks with simulated unseen entities and relations.
Furthermore, by combining the capability of entity extrapolation from subgraph encoding and relation extrapolation based on the ontology of relations, RMPI \cite{RMPI-ICDE2023} provides a nascent exemplar of this perspective.
\subsection{Temporal and Lifelong Setting}
In practical applications, some KGs include temporal constraints, which necessitates the consideration of temporal information when scoring a triple \cite{BOP2021,KGRSurvey}. Temporal KGs also face the challenge of emerging elements due to their dynamic nature. To address this issue, FILT \cite{FILT-AKBC2022} defines a problem of entity extrapolation in temporal KGs and utilizes a time-aware graph encoder and entity concept information to obtain embeddings for unseen entities. Additionally, existing knowledge extrapolation works typically assume one-time extrapolation, where all unseen elements emerge simultaneously in a batch. However, recent literature, such as ARGCN \cite{ARGCN-CIKM2022} and LKGE \cite{LKGE-AAAI2023}, considers scenarios where unseen elements emerge in a multi-batch and lifelong manner.

\section{Conclusion}

Recent years have seen an increasing amount of research on generalizing to unseen elements in KGs from diverse perspectives. In this paper, we provide a comprehensive survey of these works and summarize them under a general set of terminologies. We categorize existing methods using our proposed systematic taxonomy and list commonly used benchmarks along with the methods that employ them. We also provide method comparisons and discussions from commonly mentioned perspectives in the existing literature. Finally, we suggest several potential research directions. We believe that this exploration can provide a clear overview of the field and facilitate future research.

\clearpage

\section*{Acknowledgments}
This work is supported by the National Natural Science Foundation of China (NSFCU19B2027, NSFC91846204), the joint project DH-2022ZY0012 from Donghai Lab and the Chang Jiang Scholars Program (J2019032). Mingyang Chen is supported by the China Scholarship Council (No. 202206320309).

\section*{Contribution Statement}
Mingyang Chen and Wen Zhang contributed equally and share first authorship. 

\bibliographystyle{named}
\bibliography{ijcai23}

\end{document}

%% file: category.tex
\tikzstyle{my-box}=[
    rectangle,
    draw=hidden-draw,
    rounded corners,
    text opacity=1,
    minimum height=1.5em,
    minimum width=5em,
    inner sep=2pt,
    align=center,
    fill opacity=.5,
]
\tikzstyle{leaf}=[my-box, minimum height=1.5em,
    fill=hidden-orange!60, text=black, align=left,font=\scriptsize,
    inner xsep=2pt,
    inner ysep=4pt,
]
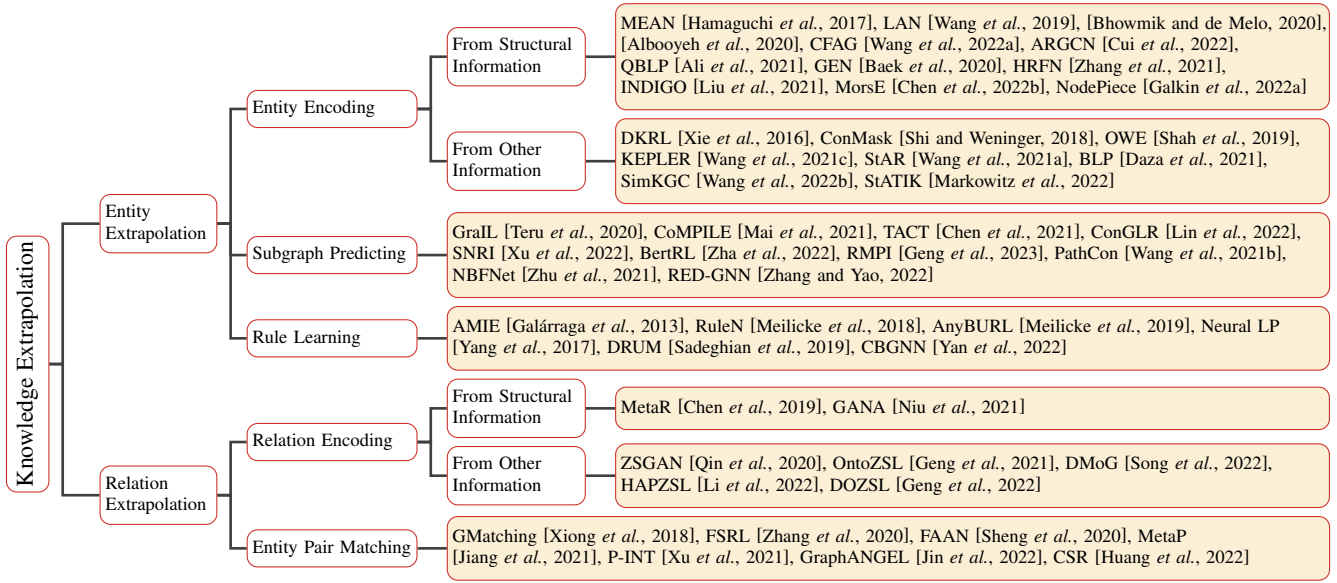
\begin{figure*}[tp]
    \centering
    \resizebox{\textwidth}{!}{
        \begin{forest}
            forked edges,
            for tree={
                grow=east,
                reversed=true,
                anchor=base west,
                parent anchor=east,
                child anchor=west,
                base=left,
                font=\small,
                rectangle,
                draw=hidden-draw,
                rounded corners,
                align=left,
                minimum width=4em,
                edge+={darkgray, line width=1pt},
                s sep=3pt,
                inner xsep=2pt,
                inner ysep=3pt,
                ver/.style={rotate=90, child anchor=north, parent anchor=south, anchor=center},
            },
            where level=1{text width=4em,font=\scriptsize,}{},
            where level=2{text width=6em,font=\scriptsize,}{},
            where level=3{text width=4.8em,font=\scriptsize,}{},
            where level=4{text width=6.1em,font=\scriptsize,}{},
            [
                Knowledge Extrapolation, ver
                [
                    Entity \\ Extrapolation
                    [
                        Entity Encoding
                        [
                            From Structural \\ Information
                            [
                                MEAN \cite{KnowTransOOKB-IJCAI2017}{,}
                                LAN \cite{LAN-AAAI2019}{,}
                                \cite{Explainable-ISWC2020}{,} \\
                                \cite{OOS-EMNLP2020}{,}
                                CFAG \cite{CFAG-AAAI2022}{,}
                                ARGCN \cite{ARGCN-CIKM2022}{,} \\
                                QBLP \cite{QBLP-ISWC2021}{,}
                                GEN \cite{GEN-NIPS2020}{,} 
                                HRFN \cite{HRFN-CIKM2021}{,} \\
                                INDIGO \cite{INDIGO-NIPS2021}{,}
                                MorsE \cite{MorsE-SIGIR2022}{,}
                                NodePiece \cite{NodePiece-ICLR2022}
                                , leaf, text width=26.5em
                            ]
                        ]
                        [
                            From Other \\ Information
                            [
                                DKRL \cite{DKRL-AAAI2016}{,}
                                ConMask \cite{ConMask-AAAI2018}{,}
                                OWE \cite{OWE-AAAI2019}{,} \\
                                KEPLER \cite{KEPLER-TACL2019}{,}
                                StAR \cite{StAR-WWW2021}{,}
                                BLP \cite{BLP-WWW2021}{,} \\
                                SimKGC \cite{SimKGC-ACL2022}{,} 
                                StATIK \cite{StATIK-NAACL2022}
                                , leaf, text width=26.5em
                            ]
                        ]
                    ]
                    [
                        Subgraph Predicting
                        [
                                GraIL \cite{GraIL-ICML2020}{,}
                                CoMPILE \cite{CoMPILE-AAAI2021}{,}
                                TACT \cite{TACT-AAAI2021}{,} 
                                ConGLR \cite{ConGLR-SIGIR2022}{,} \\
                                SNRI \cite{SNRI-IJCAI2022}{,} 
                                BertRL \cite{BertRL-AAAI2022}{,} 
                                RMPI \cite{RMPI-ICDE2023}{,}
                                PathCon \cite{PathCon-KDD2021}{,} \\
                                NBFNet \cite{NBFNet-NIPS2021}{,}
                                RED-GNN \cite{RED-GNN-WWW2022}
                                , leaf, text width=32.85em
                        ]
                    ]
                    [
                        Rule Learning
                        [
                            AMIE \cite{AMIE-WWW2013}{,}
                            RuleN \cite{RuleN-ISWC2018}{,}
                            AnyBURL \cite{AnyBURL-IJCAI2019}{,}
                            Neural LP \\ \cite{NeuralLP-NIPS2017}{,}
                            DRUM \cite{DRUM-NIPS2018}{,}
                            CBGNN \cite{CBGNN-ICML2022}
                            , leaf, text width=32.85em
                        ]
                    ]
                ]
                [
                    Relation \\ Extrapolation
                    [
                        Relation Encoding
                        [   
                            From Structural \\ Information
                            [
                                MetaR \cite{MetaR-EMNLP2019}{,}
                                GANA \cite{GANA-SIGIR2021}
                                , leaf, text width=26.5em
                            ]
                        ]
                        [
                            From Other \\ Information
                            [
                                ZSGAN \cite{ZSGAN-AAAI2020}{,}
                                OntoZSL \cite{OntoZSL-WWW2021}{,}
                                DMoG \cite{DMoG-COLING2022}{,} \\
                                HAPZSL \cite{HAPZSL-NeuroComp2022}{,}
                                DOZSL \cite{DOZSL-KDD2022}
                                , leaf, text width=26.5em
                            ]
                        ]
                    ]
                    [
                        Entity Pair Matching
                        [
                            GMatching \cite{GMatching-EMNLP2018}{,}
                            FSRL \cite{FSRL-AAAI2020}{,}
                            FAAN \cite{FAAN-EMNLP2020}{,}
                            MetaP \\ \cite{MetaP-SIGIR2021}{,} 
                            P-INT \cite{P-INT-EMNLP2021}{,}
                            GraphANGEL \cite{GraphANGEL-ICLR2022}{,}
                            CSR \cite{CSR-NIPS2022}
                            , leaf, text width=32.85em
                        ]
                    ]
                ]
            ]
        \end{forest}
    }
    \caption{Taxonomy of knowledge extrapolation for knowledge graphs.}
    \label{fig:texonomy}
\end{figure*}